\documentclass{article}

\usepackage{arxiv}

\usepackage[utf8]{inputenc} 
\usepackage[T1]{fontenc}    
\usepackage{hyperref}       
\usepackage{url}            
\usepackage{booktabs}       
\usepackage{amsfonts}       
\usepackage{nicefrac}       
\usepackage{microtype}      
\usepackage{lipsum}		
\usepackage{graphicx}
\usepackage{natbib}
\usepackage{doi}

\AtBeginDocument{%
	\providecommand\BibTeX{{%
			\normalfont B\kern-0.5em{\scshape i\kern-0.25em b}\kern-0.8em\TeX}}}

\usepackage{hyperref}
\usepackage{url}
\usepackage{graphicx}
\usepackage[ruled,vlined]{algorithm2e}
\usepackage[utf8]{inputenc} 
\usepackage[T1]{fontenc}    
\usepackage{hyperref}       
\usepackage{url}            
\usepackage{booktabs}       
\usepackage{amsfonts}       
\usepackage{nicefrac}       
\usepackage{microtype}      
\usepackage{tabularx} 
\usepackage{adjustbox}
\usepackage{color, colortbl}
\usepackage[utf8]{inputenc} 
\usepackage[T1]{fontenc}    
\usepackage{hyperref}       
\usepackage{url}            
\usepackage{booktabs}       
\usepackage{amsfonts}       
\usepackage{nicefrac}       
\usepackage{microtype}      
\usepackage{graphicx}
\usepackage{natbib}
\usepackage{color}
\usepackage{subcaption}
\usepackage{multirow}
\usepackage{caption}
\newcolumntype{L}{>{\raggedright\arraybackslash}X}
\makeatletter
\newcommand{\printfnsymbol}[1]{%
  \textsuperscript{\@fnsymbol{#1}}%
}

\title{MLIM: Vision-and-Language Model Pre-training with Masked Language and Image Modeling}

\author{
 Tarik Arici\thanks{These authors contributed equally.}\\
Amazon.com Inc\\
Seattle, WA, USA \\
  \texttt{aricit@amazon.com} \\
   \And
 Mehmet Saygin Seyfioglu\printfnsymbol{1}\\
Amazon.com Inc\\
Seattle, WA, USA \\
  \texttt{mseyfiog@amazon.com} \\
  \And
 Tal Neiman \\
Amazon.com Inc\\
Seattle, WA, USA \\
 \texttt{talneim@amazon.com} \\
 \And
Yi Xu \\
Amazon.com Inc\\
Seattle, WA, USA \\
 \texttt{yxaamzn@amazon.com} \\
 \And
Son Tran \\
Amazon.com Inc\\
Seattle, WA, USA \\
  \texttt{sontran@amazon.com} \\  
 \And
Trishul Chilimbi \\
Amazon.com Inc\\
Seattle, WA, USA \\
  \texttt{trishulc@amazon.com} \\  
 \And
Belinda Zeng\\
Amazon.com Inc\\
Seattle, WA, USA \\
  \texttt{zengb@amazon.com} \\  
 \And
Ismail Tutar\\
Amazon.com Inc\\
Seattle, WA, USA \\
  \texttt{ismailt@amazon.com} \\    
}
\begin{document}
\maketitle
\begin{abstract}
Vision-and-Language Pre-training (VLP) improves model performance for downstream tasks that require image and text inputs. Current VLP approaches differ on (i) model architecture (especially image embedders), (ii) loss functions, and (iii) masking policies. Image embedders are either deep models like ResNet or linear projections that directly feed image-pixels into the transformer. Typically, in addition to the Masked Language Modeling (MLM) loss, alignment-based objectives are used for cross-modality interaction, and RoI feature regression and classification tasks for Masked Image-Region Modeling (MIRM). Both alignment and MIRM objectives mostly do not have ground truth. Alignment-based objectives require pairings of image and text and heuristic objective functions. MIRM relies on object detectors. Masking policies either do not take advantage of multi-modality or are strictly coupled with alignments generated by other models. In this paper, we present Masked Language and Image Modeling (MLIM) for VLP. MLIM uses two loss functions: Masked Language Modeling (MLM) loss and image reconstruction (RECON) loss. We propose Modality Aware Masking (MAM) to boost cross-modality interaction and take advantage of MLM and RECON losses that separately capture text and image reconstruction quality. Using MLM + RECON tasks coupled with MAM, we present a simplified VLP methodology and show that it has better downstream task performance on a proprietary e-commerce multi-modal dataset. 
\end{abstract}


\maketitle

\section{Introduction}
\label{sec:introduction}

Since the emergence of transformer architectures \citep{Vaswani2017}, and their first successful use in Natural Language Processing (NLP) tasks \citep{bert},  they have been used for vision tasks \citep{vit, pyramid}, and finally in vision-and-language tasks \citep{vlbert, vilbert, oscar}. For vision-and-language pre-training, researchers have proposed i) diverse architectures, ii) loss functions, and iii) masking policies. Model architectures employ different techniques to convert images into a suitable input sequence to the transformer. Loss functions are designed for image-region modeling and cross-modality interaction via alignment functions between text and images. Masking strategies aim to avoid disrupting this alignment. Each of these techniques has advanced the state of the art and provided valuable insights. However, it is unclear how to integrate all of them to achieve a level of simplicity and accuracy for vision-and-language pre-training that is comparable to text transformer pre-training using masked language modeling (MLM) loss and text-token masking.

Image embedders convert an image to an embedding sequence. Image embedding sequences can be concatenated with text token embedding sequences and forwarded to the transformer. Most VLP models use object detectors as image embedders. These object detectors are deep convolutional models trained on Visual Genome \citep{genome} or ImageNet classification \citep{imagenet} datasets. Object detectors are limited to their original detection categories and are heavyweight models. While end-to-end training can adapt these models, computational complexity or end-to-end depth of the VLP model (object detector + transformer) can be an issue for training or inference. Visual transformers (ViLT) remove deep image-embedders and directly feed pixel-level inputs to the transformer via a linear projection on image patches defined by a grid \citep{vilt}. While ViLT significantly reduces computational complexity, it leaves image feature learning to the transformer layers, limiting interaction between text and image embeddings to later layers of the transformer. We propose using shallow CNN models as embedders to learn high-level features for low-level pixel-data and enable better cross-modality interaction throughout the transformer layers.

In addition to MLM loss for language modeling, Masked Image-Region Modeling (MIRM) uses various loss functions for image modeling. UNITER proposes loss functions that use RoI-features and predicted object-classes obtained from the object detector as target features and target labels for masked image regions (Chen et al., 2020). Another category of loss functions is alignment-based functions to encourage cross-modality interaction. Some alignment-based objectives aim to match fused representations of images and text (e.g., Image Text Matching (ITM)). Other alignment objectives operate on a finer resolution and aim to align text or image representations at a sequence position level via transport plans. ITM tasks require positive and negative pairs. We believe that training using such pairs is a better fit for downstream tasks and negative/positive definition might hurt generalization while pre-training. In addition, these loss functions do not use ground-truth labels, and instead rely on heuristics. We aim to design a pre-training methodology that is as well-defined as MLM loss on text tokens: ground truth targets with loss functions that are not based on heuristics. We use the image reconstruction (RECON) task for Masked Image Modeling (MIM) with the image itself as the target for this task.

While text token masking is straightforward, masking for images modeling is not. Image regions detected by the object detector are good candidates for masking as they define semantically significant regions in the image. However they do not rely on ground truth, using predictions of the object detector and require heavyweight models. In addition, masking can break cross-modal alignment if it is performed independently on text and image inputs. To avoid these requirements for masking corresponding representations in both modalities, coarse alignments obtained using text generators are utilized \citep{kaleido}. We avoid these problems by not using object detectors or alignment based objectives.

In this paper, we introduce Masked Language and Image Modeling (MLIM) for VLP pe-training. We use Transformers \citep{Vaswani2017} as the main component in the model. We use a shallow-CNN based image embedder to embed images and perform masking, and a CNN based image decoder to compute RECON loss. We use MLM loss to measure text token prediction performance and RECON loss to measure image reconstruction performance. While MLM loss is only defined on masked text tokens, RECON loss is defined over all pixels of the image. Together, MLM and RECON losses capture text and image reconstruction performance. We do not use any alignment-based loss-function for cross-modality interaction and instead use Modality-Aware Masking (MAM) to achieve this. MAM has three operating modes: (1) heavy image-masking, (2) heavy text-masking, (3) light image-masking and light text-masking. The first two modes encourage information flow across modalities. Using an image embedder, we aim to bring image embeddings to a common embedding space with text token embeddings. We perform image masking on the image embedder outputs. MAM masks image and text inputs before the transformer, and MLM + RECON losses measure reconstructed text and image quality. We demonstrate the effectiveness of our VLP pre-training methodology on a proprietary e-commerce multi-modal dataset.

Our contributions are summarized as follows:
\begin{itemize}
	\item We propose using image reconstruction (RECON) loss and show pre-training with MLM + RECON obviates the need for alignment-based and image-region modeling based loss functions. This simplifies loss function design and does not require image-text pairings to create training instances.
	\item We use a shallow CNN as an image embedder. Our image embedder is much more lightweight than deep models like ResNet, and image masking friendly.  We show that a shallow CNN image embedder can bring image pixels to higher level representations that foster interaction between image and text embeddings.
	\item We show that MAM takes advantage of the two loss functions capturing text and image reconstruction quality via masking policies that encourage cross-modal information flow.
\end{itemize}

\section{Related Work}
\label{sec:relatedwork}
Recently, transformer based model pre-training has been successfully applied in NLP tasks [elmo, bert gpt2, roberta, albert, etc]. As an extension, multimodal models are proposed to learn representations from image-text pairs. Fine-tuning multimodal models achieve state--of-the-art (SoTA) on downstream tasks. Some of these models use two-stream stream architectures that use two single-modal transformers, and their outputs are fused by a third transformer [vilbert, lxmert]. On the other hand single-stream architectures input image and text embeddings into a transformer. Single-stream models have the potential to enable better cross-modality interaction since attention mechanisms can attend to both image and text. VL-BERT presents promising results for single-stream models. More work on VL tasks are proposed lately: VLP, ViLBERT, VL-BERT, ViLT offer single-stream models for generic understanding tasks such as Visual Question Answering (VQA) and image-text retrieval or generation tasks such as image captioning. 

Object detectors are commonly used as image embedders. Image embeddings are either region of interest (RoI) features obtained from the object detector (\emph{e.g.}, \citep{uniter, lxmert, oscar}) or output feature grids of the CNN backbone \citep{jiang, nguyen}. While using grid features instead of RoI-features falls behind, it achieves similar performance with heavier CNN models \citep{pixelbert}. ViLT avoids using heavyweight image embedders by directly embedding low-level pixel data with a single-layer projection and achieves similar results with reduced complexity.

Multiple loss functions are proposed for better learning from images. Some of these can be categorized under masked image-region modeling (MIRM) tasks. RoI-feature regression tasks regresses the RoI-feature of the masked region with L2 loss. RoI-classification task learns the labels of masked regions with cross-entropy loss. UNITER \citep{uniter} and LXMERT \citep{lxmert} utilize object-detector features and its label-predictions as targets in MIRM losses. Contrastive loss in the form of detecting randomly replaced RoI tags are proposed in Oscar \citep{oscar}. Object tags are also generated by the object detector. These MIRM tasks rely on object-detectors and pseudo-targets (\emph{i.e.}, RoI features and class-labels obtained from the detector), which may be wrong or irrelevant in the given context.

Alignment-based loss functions are also commonly utilized \citep{kaleido, oscar, vilt} for cross-modality interaction. These alignments either happen at modality level as in widely used Image Text Matching (ITM) task or at finer-grained levels as in Word-Region Alignment (WRA) via Optimal Transport (OT) \citep{uniter}. Alignment task requires pairings. While creating negative pairs for ITM is easy, how to retrieve these pairs at train-time and how to find good training pairs remains an issue. Also, WRA does not have ground truth.

Our key differences from prior art is as follows: i) We simplify loss function design by using MLM and RECON losses both of which have ground truths ii) We use MAM to force cross-modal information flow for better cross-modality interaction and take advantage of MLM + RECON iii) We utilize a shallow CNN as image embedder, which is specifically designed for image embedding and masking. Our image embedder can bring low-level pixel data to a common representation space with text-token embeddings for better cross-modality interaction\footnote{We speculate low-level pixel embeddings can not directly interact with higher-level text-token embeddings. This might explain why ViLT did not work with BERT as the main transformer but it worked with Vision Transformer \citep{vit} as the main transformer.}

\section{Masked Language and Image Modeling}
\label{sec:mlim}
\subsection{Model Overview}
\begin{figure*}[t]
	\centering
	\includegraphics[scale=0.7]{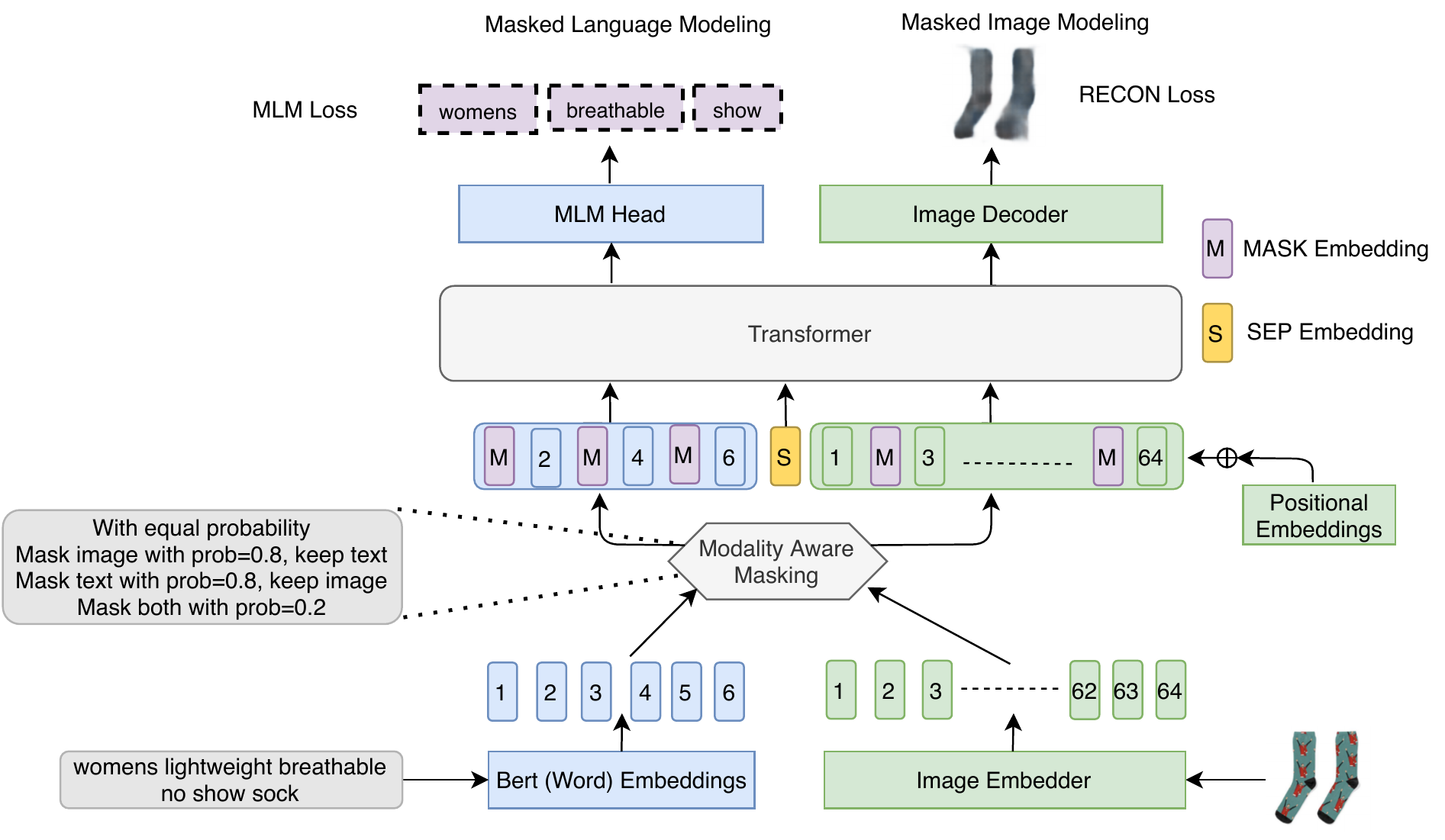}
	\caption{Model overview. Model consists of word embeddings, MLM head, CNN-based image embedder, and CNN-based image decoder, and a multi-layer self-attention Transformer as the main component. MAM operates on word and image embeddings before inputting to the Transformer.} \label{fig:arch}
\end{figure*}

The model architecture of MLIM is given in Figure~\ref{fig:arch}. Given an image-text pair, text is tokenized and embedded using transformer's word embeddings and word-position embeddings. Image is embedded using a shallow CNN model and image-positional embeddings. 

Specifically, image embedder is a CNN model with convolutional layers using 2D filters of kernel size 2x2 and stride of 2. Choosing kernel size in horizontal and vertical dimensions to be equal to the stride ensures each input will only contribute to a single filter's output. Therefore, image embeddings will correspond to non-overlapping pixel regions in the original image-resolution and masking an embedding deletes all information captured from a region in the image. Hence, prediction of masked region's pixel data will be left to transformer's self-attention layers to gather information from existing text and image embeddings. We observed that this helps with text-to-image information flow. Text is tokenized using transformer's tokenizer (\emph{e.g.} WordPiece for BERT). Word embeddings are obtained from transformer's word embeddings. Both image and word embeddings have positional embeddings separately, which are added to the embeddings. Since we have separate positional embeddings for the two modalities, it is redundant to have modality embeddings. Therefore, we do not use any explicit modality embeddings (i.e., “segment embedding” in BERT).

Image decoder inputs transformer-outputs corresponding to image embeddings. Transformer outputs (1D vector sequences) are reshaped to 2D vector grids for deconvolutional filtering.  Image decoder consists of cascaded deconvolutional layers to bring 2D vector grids back to the original image resolution with three color channels. Output of the decoder is a tensor of shape, \emph{i.e.}, (\texttt{image\_width},  \texttt{image\_height}, 3). Finally an element-wise sigmoid function is applied to the decoder output tensor to match pixel-intensity range (\emph{i.e.}, [0-1]) for each color channel.

We have two pre-training tasks: Masked Language Modeling (MLM) and image reconstruction (RECON) coupled with Modality Aware Masking (MAM). MAM applies masking to both word and image embeddings. Both image and word masking is realized by replacing an embedding with the embedding of \texttt{[MASK]}. This way transformer layers recognize \texttt{[MASK]}'s embedding as a special embedding that needs to be ``filled in'', independent of the modality, by attending to other vectors in the layer inputs\footnote{We observed that using separate mask embeddings for the two modalities as in UNITER degrades the performance.}. We do not mask positional embeddings, therefore they are added after masking. MAM operates on three modes as given in Figure~\ref{fig:arch}. To pre-train, we apply MLM + RECON as a multi-loss objective given a mini-batch of image-text pairs.

\subsection{Pre-train Objectives for MLIM}

\textbf{Masked Language Modeling (MLM)}: This task's objective is to predict the masked words from available words and image regions. We follow BERT for this task: two-layer MLP MLM head outputting logits over the vocabulary. MLM loss is negative log-likelihood for masked words. MAM determines the masking probability.

\textbf{Masked Image Modeling (MIM)}: This task's objective is to reconstruct the full image from available words and image regions. Our RECON loss is an average of pixel-wise sum of squared errors (SSE).

Both tasks aim to \emph{reconstruct} masked image and text inputs. While MLM task only reconstructs masked tokens, MIM task reconstructs the full image. This is because image embedder outputs embeddings corresponding to input image regions which are \emph{lossy} representations of the pixel data. Hence MIM task is better defined as reconstruction of masked and not-masked image regions, \emph{i.e.}, the whole image.

We do not have any other tasks, specifically no image-region feature prediction, image-region classification, or any alignment loss at modality or embedding sequence-position level.
\section{Experiments}
\label{sec:experiments}
We use Amazon catalog data for pre-training. Amazon catalog includes items available online. These items have images and textual attributes. Catalog item images are dominantly single-item images and the text describes attributes of the item. We sampled 6M items with their relevant attributes from our catalog for pre-training.

We evaluate MLIM on an Amazon internal dataset collected for the task of finding closely-matching (CM) item pairs. A pair of items is labeled as a match or a mismatch depending on the type of variation between the two items. Learning and predicting relationships is an essential task in our catalog systems and requires both image and text as input. CM dataset has 30K training examples and 10K test examples.

\subsection{Implementation Details}
For all experiments we use Adam optimizer with a fixed learning rate of $8*10^{-4}$.  We resize images to $384\times384$ resolution. We use pre-trained BERT model (\texttt{bert-large-uncased}) from Huggingface \citep{huggingface} with 24 stacked transformer blocks, 16 attention heads, and 1024 hidden state dimension as the transformer and continue training on Amazon catalog pre-training dataset. \texttt{bert-large-uncased} model has 336M parameters in total. Our image embedder and decoders are randomly initialized. Image embedder is a CNN with 7 layers and 200K parameters. Image decoder is a deconvolutional network with 10 layers and 2.8M parameters. Encoder output is a grid of $8\times8$, which is reshaped to a 1D sequence of length 64 for inputting to the transformer. We intentionally kept the encoder lightweight and decoder heavyweight.

\subsection{Results}

\begin{table}[t]
	\caption{PR AUC values on CM test dataset: Modality ablation study and the effect of MDO on fine-tuning}
	\label{tab:mdo}
	\centering
	\begin{tabular}{ccc}
		\toprule
		Model & 	Description &  PR AUC\\
		\midrule
		Image and text with MDO & Pre-trained with RECON, MLM and MAM, fine-tuned with MDO &0.884\\
		Image and text without MDO & Pre-trained with RECON, MLM and MAM, fine-tuned without MDO &0.864\\
	\end{tabular}
\end{table}

\subsubsection{Fine-tuning for Pairwise Downstream Tasks}
We fine-tune and evaluate our model on the CM downstream task. A pair of image-text inputs are concatenated using the separator token. We remove the image decoder from the model-graph and do not apply any masking strategy for fine-tuning.

Although fine-tuning VLP models on downstream tasks is mostly straightforward, VLP models offer new fine-tuning tricks. Since our CM task requires a pair of image+text inputs, we exploit this multi-modality by employing Modality Dropout (MDO). MDO improves fine-tuning by randomly dropping one of the modalities. Similar to MAM, MDO operates in one of the three modes on a micro-batch: text-only, image-only, and image-text mode. In Table~\ref{tab:mdo}, we present PR AUC values. Fine-tuning with MDO further improves performance.

We present an ablation study for loss functions in Table~\ref{tab:loss}. Using RECON loss instead of ITM loss improves PR AUC from 0.855 to 0.884. Using ITM loss together with MLM and RECON does not change the performance.

\begin{table}[t]
	\caption{PR AUC values on CM test dataset: Loss function ablation study and the effect of MAM}
	\label{tab:loss}
	\centering
	\begin{tabular}{ccc}
		\toprule
		Model & 	Description &  PR AUC\\
		\midrule
		RECON + ITM + MLM + MAM& Pre-trained with RECON, ITM, MLM and MAM &0.884\\		
		RECON + MLM + MAM& Pre-trained with RECON, MLM and MAM &0.884\\
		RECON + MLM + Naive Masking &	Pre-trained with RECON, MLM, and fixed masking prob of 0.2& 0.873\\
		ITM + MLM + MAM & Pre-trained with ITM, MLM and MAM & 0.855\\
		ITM + MLM + Naive Masking & Pre-trained with ITM, MLM, and fixed masking prob of 0.2&0.8\\
		\bottomrule
	\end{tabular}
\end{table}

\subsubsection{Cross-Modality Interaction}

\begin{figure}
     \centering
     \begin{subfigure}[b]{0.45\textwidth}
         \centering
         \includegraphics[width=\textwidth]{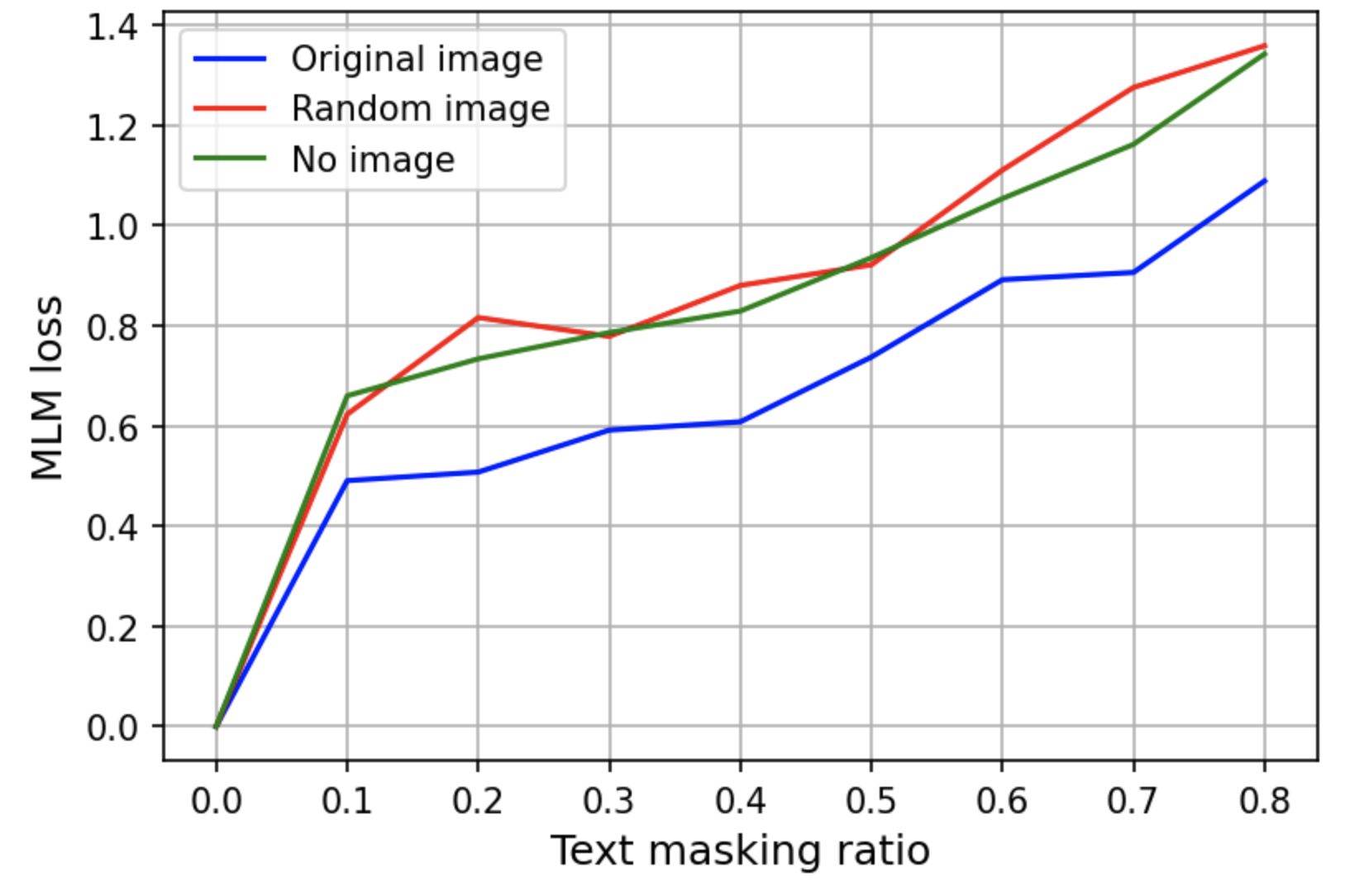}
         \caption{Effect of images on MLM loss}
         \label{fig:mlm_loss}
     \end{subfigure}
     \hfill
     \begin{subfigure}[b]{0.45\textwidth}
         \centering
         \includegraphics[width=\textwidth]{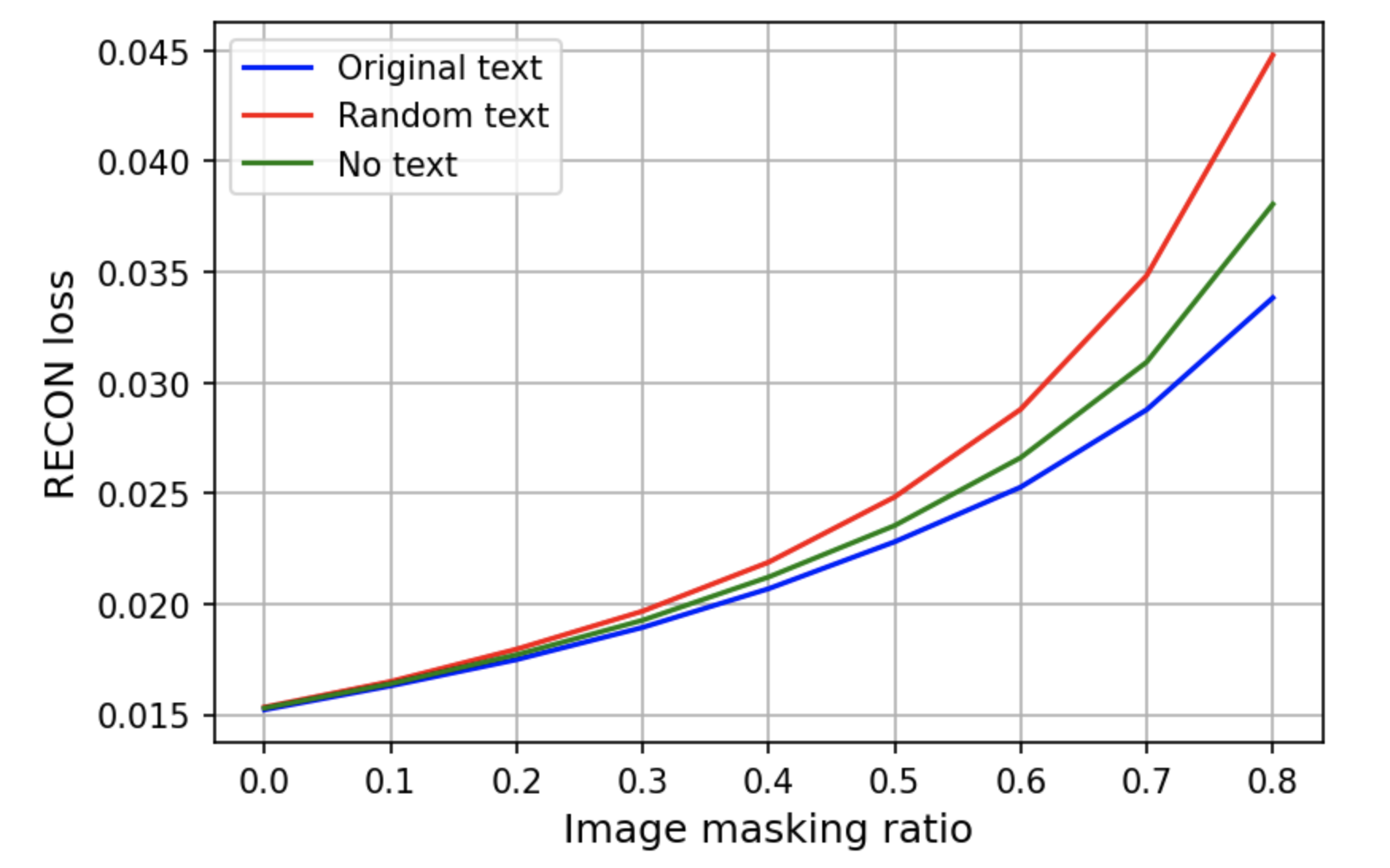}
         \caption{Effect of text on RECON loss}
         \label{fig:recon_loss}
     \end{subfigure}
        \caption{We use MLM and RECON loss on pre-training test-data to measure performance improvements by cross-modal information flow}
        \label{fig:three graphs}
\end{figure}

As discussed before, cross-modality interaction in the transformer is a desired objective in VLP models. In this section, we show evidence for cross-modality interaction in our proposed method. We present MLM and RECON losses on our test data under different inputting schemes. In Figure~\ref{fig:mlm_loss}, we show MLM loss computed on text inputs paired with random images sampled from the dataset, text inputs paired with no images (specifically gray images) and text inputs paired with their original images in the dataset for different text-token masking probabilities. MLM loss is reduced by using images, showing that the transformer uses information from the image inputs to achieve the MLM task. In Figure~\ref{fig:recon_loss}, we show RECON loss computed on image inputs paired with random texts in the dataset, image inputs paired with empty texts and image inputs paired with their original texts in the dataset for different image-embedding masking probabilities. RECON loss is reduced by using text, showing that the transformer uses information from the text inputs to achieve the RECON task. 

We note that random text inputs degrade RECON performance more than random images degrade MLM performance. This might be implying that text-to-image information flow is more significant compared to image-to-text information flow.

\section{Conclusion}
\label{sec:conclusion}
In this paper, we present Masked Language and Image Modeling (MLIM): a simplified VLP method using MLM and RECON losses and MAM. We simplify loss function design, propose a shallow image embedder to avoid heavyweight object-detectors and present an image decoder to enable RECON loss. We believe VLP datasets (\emph{e.g.} e-commerce datasets) are large enough to enable learning built-in image embedders during pre-training. 

While alignment-based loss functions are promising and help in learning contrastive features, finding good image-text pairs (especially negative pairs) becomes an issue and makes pre-training rely on pairing techniques. On the other hand finer-grained alignment objectives do not have ground truth.

Masked Image-Region Modeling (MIRM) relies on RoI features and classes predicted by the object detector. Furthermore MIRM tasks aim to ``fill in'' masked regions. However RECON task aims to reconstruct the whole image.

We encourage work in built-in image embedders designed to be masking friendly, and in image decoders that are designed to get the best cross-modality interaction inside the transformer.

As a future work, we will optimize image embedder and decoder for public pre-training datasets which have more complexity (\emph{e.g.}, multiple objects and complex scenes) compared to e-commerce data.
\bibliographystyle{ACM-Reference-Format}
\bibliography{library.bib}

\end{document}